\documentclass[11pt]{article}
\usepackage[latin9]{inputenc}
\usepackage{float}
\usepackage{amsmath}
\usepackage{amssymb}
\usepackage{graphicx}
\usepackage[a4paper]{geometry}
\usepackage{setspace}
\makeatletter

\providecommand{\tabularnewline}{\\}

\numberwithin{equation}{section}
\newenvironment{lyxlist}[1]
	{\begin{list}{}
		{\settowidth{\labelwidth}{#1}
		 \setlength{\leftmargin}{\labelwidth}
		 \addtolength{\leftmargin}{\labelsep}
		 }}
	{\end{list}}

\usepackage{paralist}
 
\input{tcilatex}
\renewcommand{\thesection}{\arabic{section}.}
\usepackage{titlesec}
\titleformat{\section}
  {\normalfont\bfseries\centering}{\thesection}{0.3em}{}

\makeatother

\begin{document}
\title{\textbf{Collective intelligence through aggregation\thanks{We thank two anonymous referees and an editor for very helpful comments.
Published version in \emph{Philosophical Transactions of the Royal
Society B} (2026) 381 (1948): 20240454, available at https://doi.org/10.1098/rstb.2024.0454. }}\vspace{-0.1cm}
}
\author{{\Large Franz Dietrich and Christian List}}
\date{{\small\vspace{-1cm}
}}

\maketitle
\noindent\noindent\rule{\linewidth}{0.4pt}

\noindent Suppose a committee, expert panel, or other group is making
judgments on some issues, where these may be not just yes/no-questions,
such as whether a defendant is guilty, but also variables with many
possible values, such as macroeconomic or meteorological variables
or travel directions. Furthermore, there may be interconnections between
different issues, as in the case of economic or climate variables.
How can the group arrive at \textquotedblleft intelligent\textquotedblright{}
collective judgments, based on the group members\textquoteright{}
individual judgments? We investigate three challenges raised by this
judgment-aggregation problem. First, reasonable methods of aggregation
(such as defining the collective judgment for each issue as the average
or median judgment) can produce inconsistent collective judgments.
Secondly, many methods of aggregation are manipulable by strategic
voting. Finally, not all methods of aggregation are conducive to tracking
the truth on the issues in question. We prove new impossibility or
possibility theorems on all three challenges, identifying what it
takes to produce collective judgments in a consistent, non-manipulable,
and truth-tracking manner and thereby to achieve collective intelligence
through aggregation. Overall, the median method, though imperfect,
performs reasonably well. We also note the relevance of our analysis
for non-human group decisions. 

\noindent\noindent\rule{\linewidth}{0.4pt}

\section{Introduction}

In a much-cited experiment reported in \emph{Nature} in 1907 \cite{Galton},
the statistician Francis Galton asked the participants of an agricultural
fair in rural England to estimate the weight of an ox on display.
Around 800 people submitted their estimates. Although the participants
were familiar with farm animals, the estimates varied significantly.
But the \emph{median} estimate \textendash{} the one exactly in the
middle when all submitted estimates were listed in ascending order
\textendash{} came remarkably close to the true weight. The true weight
was 1198lbs, and the median estimate was 1207lbs, an error of only
0.75\%. The average individual error was far greater. Galton's experiment
illustrates the idea of ``collective intelligence through aggregation'':
asking many individuals to express a judgment on some issue and deriving
from the answers a collective judgment \textendash{} for short, \emph{aggregating
the individual judgments} \textendash{} can be a better way to ``track
the truth'' than to ask a single individual alone \cite{ElsterLandemore}.
Moreover, this can work even if there is significant disagreement
among the participants. 

In Galton\textquoteright s study, the method of aggregation was the
median rule \textendash{} taking the median individual judgment as
the collective one \textendash{} but different methods might be used
in other cases, such as averaging. Galton\textquoteright s example
is an instance of what we call a \textquotedblleft general judgment
aggregation problem\textquotedblright :
\begin{itemize}
\item A group of individuals \textendash{} e.g., a committee, an expert
panel, or even a non-human group \textendash{} is making judgments
(``decisions'') on one or more issues or variables, and the goal
is to arrive at a collective judgment (``decision'') on those issues
or variables; 
\item the issues or variables could be either binary (such as ``guilt''
versus ``innocence'' in a criminal jury) or many-valued (such as
the weight of an ox); and
\item when there are two or more issues or variables, these may be interconnected,
in the sense that the judgments (``decisions'') on some of them
constrain those that can consistently be made on others. 
\end{itemize}
Galton\textquoteright s example involves just a single, many-valued
issue. Many decision-making groups, however, need to estimate several
variables with constraints between them. Think of an expert panel
predicting future macroeconomic variables, such as size of the workforce,
each worker's productivity, and GDP, where the last variable is the
product of the first two. Or think of an expert panel making judgments
about a complex system involving numerous interconnected variables,
such as the Earth's climate system. 

Beyond human societies, animal groups sometimes aggregate judgments
too, for instance when they choose a travel direction, grazing spot,
or nest site from amongst several options, and here, too, the advantages
of democratic decision methods have been noted \cite{ConradtList2009,ConradtRoper2003,CouzinEtAl2005,Prins1996}.
Finally, judgment aggregation problems also arise in the context of
artificial intelligence, for instance when several data bases need
to be merged or the inputs of several AI agents need to be combined.
Table 1 provides an illustrative taxonomy of different kinds of aggregation
problems (see below for citations of relevant literature on different
such problems).
\begin{table}[H]
\caption{Examples of general aggregation problems}

\centering{}%
\begin{tabular}{|c|c|c|}
\hline 
 & One issue/variable & Many issues/variables\tabularnewline
\hline 
\hline 
\begin{tabular}{c}
Binary\tabularnewline
judgments\tabularnewline
\end{tabular} & %
\begin{tabular}{c}
Jury decisions\tabularnewline
(``guilty'', ``not guilty'')\tabularnewline
\end{tabular} & %
\begin{tabular}{c}
Making yes/no judgments on\tabularnewline
logically related propositions\tabularnewline
\end{tabular}\tabularnewline
\hline 
\begin{tabular}{c}
Non-binary\tabularnewline
judgments\tabularnewline
\end{tabular} & %
\begin{tabular}{c}
Evaluating or estimating\tabularnewline
one issue/variable\tabularnewline
Choosing a grazing spot,\tabularnewline
nest site, or travel direction\tabularnewline
\end{tabular} & %
\begin{tabular}{c}
Estimating interconnected\tabularnewline
economic or climate variables\tabularnewline
Assigning probabilities\tabularnewline
or likelihoods to several events\tabularnewline
\end{tabular}\tabularnewline
\hline 
\end{tabular}
\end{table}

\noindent In this paper, we will model such judgment aggregation problems
in full generality and investigate three challenges they may raise.
First, reasonable methods of aggregation can produce inconsistent
collective judgments. Second, many methods of aggregation are manipulable
by strategic voting. And third, only some methods successfully track
the truth. We will discuss how one might respond to those challenges.
Interestingly, the median method, as in Galton's study, performs better
overall than averaging, but it is still threatened by some of the
identified challenges.

Our analysis builds on several different literatures: the literature
on binary and non-binary judgment aggregation and social-choice theory
\cite{AczelWagner,Arrow,Basili,Claussen2010,DietrichListArrow,DokowHolzmanBinary,DokowHolzmanNonbinary,DuddyPigginsManyValued,KornhauserSager1986,ListPettit2002,NehringPuppe2010,RubinsteinFishburn,Sen1977,Wilson},
the literature on Condorcet's jury theorem \cite{DietrichList2004,GrofmanCJT,ListGoodin2001,BovensRabinowicz,DietrichSpiekermann2013,DietrichSpiekermannSEP,Pivato,Kilinc2025},
and the literature on grade aggregation \cite{BalinskiLarakiPNAS,BalinskiLaraki,Morreau2021,Kilinc2025}.
All formal proofs and some additional analyses are included in online
supplementary material (linked at https://doi.org/10.1098/rstb.2024.0454).

\section{The single-issue case}

We begin by assuming that a group makes a judgment on just a single
issue or variable. Formally, we have a set $N=\left\{ 1,2,...,n\right\} $
of two or more individuals (decision-makers, experts, judges, \dots )
and an issue or variable, denoted $x$, which takes values in some
set $V$ (e.g., the set $\mathbf{R}$ of real numbers). The goal is
to assign a value from that set to the given issue/variable (in effect,
to ``choose'' a value for $x$). We write $J_{i}$ to denote individual
$i$\textquoteright s \emph{evaluation} of that issue/variable, where
$J_{i}\in V$. This expresses $i$\textquoteright s judgment as to
which value of $x$ is the right one. In Galton\textquoteright s study,
$J_{i}$ is the weight estimate submitted by the $i^{\text{th}}$
participant. A list ($n$-tuple) of evaluations across the individuals,
written $\left\langle J_{1},...,J_{n}\right\rangle $, is called a
\emph{profile of evaluations}. Examples of issues or variables are: 
\begin{itemize}
\item the weight of an ox or another real-valued variable, say in economics
or climate science, such as GDP or the atmospheric greenhouse gas
concentration ($V=\mathbf{R}$);
\item the likelihood of an event, such as an earthquake or the melting of
a glacier, where this could be either qualitative ($V=\left\{ \text{very unlikely},\text{unlikely},\text{as likely as not},...\right\} $)
or quantitative (i.e., a probability in $V=[0,1]$); 
\item the grade of an exam paper or an athlete (e.g., $V=\left\{ \text{A},\text{B},\text{C},...\right\} $); 
\item a candidate to be chosen for some political office ($V=\left\{ \text{Harris},\text{Trump},...\right\} $);
or
\item a travel direction (e.g., $V=\left[0{^\circ},360{^\circ}\right]$),
a grazing spot, or a nest site (e.g., $V$ is a set of locations).
\end{itemize}
An \emph{aggregation rule} is a function, denoted $F$, that maps
each profile $\left\langle J_{1},...,J_{n}\right\rangle $ in some
domain of admissible profiles (e.g., all logically possible profiles)
to a collective evaluation $J=F(J_{1},...,J_{n})$, where $J$ stands
for the group's evaluation of the issue/variable $x$, with $J\in V$.
Examples of aggregation rules are the following: 
\begin{itemize}
\item \emph{Plurality rule} (or \emph{mode rule}): The collectively assigned
(``chosen'') value for issue/variable $x$ is the value most often
assigned by the individuals (and if two or more values are tied, it
is one of those values, selected by some tie-breaking criterion).
\item \emph{Median rule}: The collectively assigned value is the median
value assigned by the individuals, i.e., the value exactly in the
middle (``middlemost'') when all evaluations are listed in ascending
order (with repetitions whenever the same value is assigned more than
once), assuming an odd number of individuals.\footnote{When $n$ is even, there is a \emph{median pair} of values (e.g.,
if $n=4$, the second and third values in the ascending order). If
these coincide, we get a unique value. If not, we can extend the definition
of the median as follows: we could either (i) apply a tie-breaking
criterion (e.g., choose the lower of the two values), which would
give us a non-neutral rule; or (ii) deem all values in the interval
weakly between the lower and the higher value admissible, which would
lead to a multi-function generating a set of outputs; or (iii) randomize
between the two values, which would yield a non-deterministic rule.}
\item \emph{Quantile rule with parameter} $p$ (where $0\leq p\leq1$):
The collectively assigned value is the $p$-quantile value assigned
by the individuals, i.e., the value at position $m=\left\lceil pn\right\rceil $
in the ascending order of evaluations, where $\left\lceil pn\right\rceil $
is the smallest number in $\left\{ 1,2,...,n\right\} $ greater than
or equal to $pn$. For instance, the 0-quantile rule selects the lowest
value assigned by the individuals; the 1-quantile rule selects the
highest value; and the 0.5-quantile rule selects the median value
(assuming $n$ is odd).\footnote{Quantile rules (sometimes also called ``order functions'') have
been discussed in detail in \cite{BalinskiLarakiPNAS,Kilinc2025}.}
\item \emph{Averaging rule}: The collectively assigned value is the arithmetic
average of the values assigned by the individuals. 
\item \emph{Distance-based rule}: The collectively assigned value is a value
$J$ that minimizes the sum-total distance from the individual values
$\sum_{i=1}^{n}d(J,J_{i}$), where $d$ is some distance function
on the set $V$. If $V$ is an interval of real numbers (possibly
all of $\mathbf{R}$) the distance between any two values $J,J'$
could be defined as $d(J,J')=\vert J-J'\vert^{q}$ for some fixed
$q\geq1$. For $q=1$, the corresponding distance-based rule becomes
the median rule (assuming $n$ is odd), and for $q=2$, it becomes
the averaging rule. For other values of $q$, it is distinct.
\item \emph{Dictatorship}: The collectively assigned value is the value
assigned by some antecedently fixed individual, where the same individual
determines the collectively assigned value for every profile. 
\end{itemize}
Although, in human groups, it is natural to think of an aggregation
rule as a formal or explicit mechanism for generating collective judgments
based on individual judgments (``explicit voting''), the concept
of an aggregation rule, understood as a function from individual-level
inputs to collective-level outputs, can also accommodate informal
or implicit mechanisms (``implicit voting'' or ``voting with one's
feet''), as in the example of animal groups making collective decisions
in a participatory manner \cite{ConradtList2009}. 

What are the properties \textendash{} especially the advantages and
disadvantages \textendash{} of different aggregation rules? The first
point to note is that different rules rely on different assumptions
about the nature of the issue/variable to be evaluated. The plurality
and dictatorial rules are always formally well-defined, irrespective
of whether we are estimating a real-valued variable (such as the weight
of an ox), assigning a qualitative grade to an exam paper (such as
A, B, C etc.), choosing one of several electoral candidates (such
as Harris or Trump), or choosing a nest-site or grazing spot. Those
rules do not require any particular structure of the value set; we
can think of its elements simply as options among which a choice is
to be made. The median, quantile, and averaging rules, by contrast,
require a value set with a certain structure. For the median and quantile
rules, the set of possible values must be linearly ordered from lowest
to highest, and for the average, we need an arithmetic scale (e.g.,
the real numbers), with a value set that is closed under averaging.
This makes the averaging rule inapplicable, for instance, to grade
assignments such as A, B, C etc. or to qualitative probability assignments
such as ``unlikely'', ``likely'' etc., where the average is undefined.
The median and quantile rules can still be applied here, since the
values can be linearly ordered. A distance-based rule, finally, requires
a value set for which a distance function is defined.

Moreover, although averaging seems a particularly obvious method of
aggregation, it is meaningful only if the scale is sufficiently non-arbitrary.
If it were arbitrary whether to measure the severity of an earthquake
on a linear scale or on a logarithmic one, then averaging experts'
estimates of earthquake severity would have no meaning independently
of the chosen scale. If, on the other hand, the scale is unique up
to positive linear transformations of the form $ax+b$, with $a>0$,
then averaging does have a scale-independent meaning (technically,
averaging commutes with rescaling). Temperature is a good example.
Suppose Celsius, Kelvin, and Fahrenheit are the only admissible scales.
Then averaging several temperature estimates in Celsius and converting
the result into Fahrenheit or Kelvin yields the same aggregate estimate
as first converting the estimates into Fahrenheit or Kelvin and then
averaging them. The median and quantile rules, by contrast, do not
require an arithmetic scale. An ordering of the values suffices (on
the significance of scales, see, e.g., \cite{Stevens1946} and \cite{Sen1977}).

The median rule further admits a majoritarian rationalization \cite{BalinskiLarakiPNAS,BalinskiLaraki}:
the median value (where $n$ is odd) is the unique value for which
a majority judges that it is not too high \emph{and} a majority judges
that it is not too low. No other value has this ``democratic'' property.
Finally, the median rule is robust to outlier judgments. Had a few
participants in Galton's study erroneously estimated the weight of
the ox to be zero or a million lbs, this would not have changed the
median. The averaging rule lacks this robustness. Quantile rules can
be rationalized if we wish to find a value that only a minority considers
too high (if $p<0.5$) or only a minority considers too low (if $p>0.5$).

\section{The challenge of inconsistency}

To see how the aggregation of judgments becomes more complicated when
there are multiple issues or variables, such as in expert panels or
policy-making comittees, consider a simple example (for earlier related
examples, see \cite{Claussen2005,Claussen2010}). Suppose a three-member
expert panel seeks to estimate three macroeconomic variables for a
medium-sized country for some future year. 
\begin{lyxlist}{00.000}
\item [{$x$:}] workforce productivity, i.e., GDP per worker;
\item [{$y$:}] employment level, i.e., number of workers;
\item [{z:}] GDP.
\end{lyxlist}
It is agreed that $z=x\times y$. Suppose the experts' estimates are
as shown in the first three rows of Table 2.
\begin{table}[H]
\caption{Experts' estimates}

\centering{}%
\begin{tabular}{|c|c|c|c|}
\hline 
 & %
\begin{tabular}{c}
Workforce productivity ($x$)\tabularnewline
(EUR)\tabularnewline
\end{tabular} & %
\begin{tabular}{c}
Employment level ($y$)\tabularnewline
(number of workers)\tabularnewline
\end{tabular} & %
\begin{tabular}{c}
GDP ($z$)\tabularnewline
(EUR)\tabularnewline
\end{tabular}\tabularnewline
\hline 
\hline 
Expert 1 & 116 thousand & 25 million & 2900 billion\tabularnewline
\hline 
Expert 2 & 87 thousand & 39 million & 3393 billion\tabularnewline
\hline 
Expert 3 & 106 thousand & 26 million & 2756 billion\tabularnewline
\hline 
Median & 106 thousand & 26 million & 2900 billion\tabularnewline
\hline 
Average & 103 thousand & 30 million & $\approx$3016 billion\tabularnewline
\hline 
\end{tabular}
\end{table}
 Although the experts differ in their estimates for $x$, $y$, and
$z$, their individual estimates are consistent with the agreed constraint
$z=x\times y$. Strikingly, however, if the expert panel aggregates
these estimates via the median or the averaging rule, it reaches collective
estimates that are inconsistent with the agreed constraint. The median
estimates for $x$ and $y$ would imply a value of 2756 billion for
$z$, while the median estimate for $z$ is 2900 billion. The average
estimates for $x$ and $y$ would imply a value of 3090 billion for
$z$, while the average estimate for $z$ is $\approx$3016 billion.
So, although the median rule or the averaging rule can be applied
to each variable \emph{in isolation}, the simultaneous application
of one of these rules to \emph{all} variables yields an inconsistent
result \cite{Claussen2005,Claussen2010}. 

The observation that the aggregation of judgments via the median or
average may lead to inconsistencies resembles a classic problem concerning
the aggregation of binary (true/false) judgments, where majority voting
does not guarantee consistent collective judgments even when all individual
judgments are consistent \cite{KornhauserSager1986,ListPettit2002,ListPuppeSurvey}.
To illustrate, suppose three experts make judgments on three propositions
about the year 2050:
\begin{lyxlist}{00.000}
\item [{$p$:}] The atmospheric greenhouse gas concentration will exceed
650 ppm CO$_{2}$e.
\item [{$p\rightarrow q$:}] If the atmospheric greenhouse gas concentration
exceeds 650 ppm CO$_{2}$e, then the Artic ocean will be ice-free.
\item [{$q$:}] The Artic ocean will be ice-free.
\end{lyxlist}
If individual judgments are as shown in Table 3, for instance, then
majorities judge that $p$ is true, that $p\rightarrow q$ is true,
and yet that $q$ is false, an inconsistent set of judgments.
\begin{table}[H]
\caption{Experts' judgments}

\centering{}%
\begin{tabular}{|c|c|c|c|}
\hline 
 & $p$ & $p\rightarrow q$ & $q$\tabularnewline
\hline 
\hline 
Expert 1 & True & True & True\tabularnewline
\hline 
Expert 2 & True & False & False\tabularnewline
\hline 
Expert 3 & False & True & False\tabularnewline
\hline 
Majority & True & True & False\tabularnewline
\hline 
\end{tabular}
\end{table}

Is the failure of plausible aggregation rules to guarantee consistent
collective judgments an artifact of the averaging or median rules,
or does this problem occur more generally? To address this question,
we introduce a general model (earlier related modelling strategies
can be found in \cite{Claussen2005,Claussen2010,DietrichListPA,DokowHolzmanNonbinary,DuddyPigginsManyValued}).

As before, we have a set $N=\left\{ 1,2,...,n\right\} $ of two or
more individuals. Now there is a finite and non-empty set $X=\left\{ x,y,z,...\right\} $
of issues or variables. Again, these take values in some set $V$
(e.g., the set $\mathbf{R}$ of real numbers). Each individual $i$'s
judgments are represented by an \emph{evaluation }$J_{i}$, formally
a function from the set $X$ into the set $V$, which assigns to each
issue or variable $\text{\ensuremath{x}}$ in $X$ a value $J_{i}(x)$
in $V$, expressing $i$'s judgment about the value of $x$. 

To represent the constraints between different issues or variables,
we introduce a non-empty set $\mathcal{J}$ of evaluations (functions
from $X$ into $V$) which count as \emph{consistent}. In the example
of the three macroeconomic variables, the set $\mathcal{J}$ consists
of those evaluations $J$ which satisfy the constraint $J(z)=J(x)\times J(y)$.
Valuations breaching that constraint are excluded from $\mathcal{J}$.
If, in a more complicated case, the constraints between different
issues or variables were given by an entire system of equations or
inequalities, then the set $\mathcal{J}$ would consist of those evaluations
$J$ that respect all those equations or inequalities. In general,
the set of issues or variables $X$ together with the set $V$ of
possible values and the set $\mathcal{J}$ of consistent evaluations
represents the \emph{evaluation problem} under consideration. 

To see that this model can capture not only disagreements about the
values of the variables but also disagreements about the constraints
between the variables, note that one could include in $X$ a further
variable $c$ (for ``constraint''), which takes the value $c=1$
if the constraint between $x$, $y$, and $z$ is $z=x\times y$ and
the value $c=2$ if that constraint is, for example, $z=x+y$. An
evaluation $J$ (a function from $X$ into $V$) would then count
as consistent (and thereby be included in $\mathcal{J}$) if 
\begin{lyxlist}{00.00.0000}
\item [{\emph{either}}] $c=1$ and $J(z)=J(x)\times J(y)$ 
\item [{\emph{or}}] $c=2$ and $J(z)=J(x)+J(y)$.
\end{lyxlist}
This can be generalized to capture any possible disagreements about
the constraints between different variables.

As before, an $n$-tuple of evaluations across the individuals, $\left\langle J_{1},...,J_{n}\right\rangle $,
is called a \emph{profile of evaluations}, and an \emph{aggregation
rule} is a function, $F$, that maps each profile $\left\langle J_{1},...,J_{n}\right\rangle $
in some domain of admissible profiles to a collective evaluation $J=F(J_{1},...,J_{n})$.
For any issue or variable $x\in X$, $J(x)$ denotes the group's evaluation
of that issue/variable $x$. We have already mentioned several examples
of aggregation rules, such as the plurality, averaging, and median
rules etc. These can clearly be extended from one issue or variable
to many. The plurality rule in fact admits two distinct extensions.
One could either define the \emph{collective value for each issue/variable}
as the value most often assigned to it by the individuals (the \emph{issuewise
plurality rule}). Or one could define the \emph{collective evaluation
in its entirety} as the evaluation that is submitted most often (the
\emph{global plurality rule}).\footnote{In each case, one might need a tie-breaking criterion.}
Moreover, it may make sense to define an aggregation rule that uses
different criteria for producing the collective judgments for different
issues or variables, for instance averaging for issues such as $x$
(productivity) and $y$ (employment) in the macroeconomic example
and the plurality criterion for issues such as $c$ (the relevant
constraint). For a distance-based rule, finally, one must introduce
a distance function on the set $\mathcal{J}$ of consistent evaluations;
for each profile, the collective evaluation is then one that minimizes
the sum-total distance from the individual evaluations.\footnote{For each profile $\left\langle J_{1},...,J_{n}\right\rangle $, the
collective evaluation $J$ must be chosen to minimize $\sum_{i=1}^{n}d(J,J_{i}$).
To be \emph{well-behaved}, a distance function $d:\mathcal{J}\times\mathcal{J}\rightarrow\left[0,\infty\right)$
must satisfy (i) for all $J,J',J''$, $d(J,J')+d(J',J'')\geq d(J,J'')$,
(ii) for all $J,J'$, $d(J,J')=0\Leftrightarrow J=J'$, and (iii)
for all $J,J'$, $d(J,J')=d(J',J)$.}

Can we find compelling aggregation rules that avoid the problem of
collective inconsistency? We will now introduce some basic desiderata
that one might expect a good aggregation rule $F$ to meet (inspired
by social-choice and judgment-aggregation theory, e.g., in \cite{Arrow,DietrichListArrow,ListPettit2002,ListPuppeSurvey,NehringPuppe2010}).
Later, one might consider relaxing some of them.\smallskip{}

\noindent\textbf{Universality:} Any profile of consistent evaluations
is admissible as input to $F$. Formally, the domain of $F$ is the
set of all possible profiles of consistent evaluations $\mathcal{J}^{n}$.\smallskip{}

\noindent\textbf{Collective rationality:} The output of $F$ is a
consistent evaluation. Formally, for any admissible profile $\left\langle J_{1},...,J_{n}\right\rangle $,
$F(J_{1},...,J_{n})\in\mathcal{J}$.\smallskip{}

\noindent\textbf{Anonymity:} $F$ treats different evaluators equally.
Formally, $F(J_{1},...,J_{n})=F(J'_{1},...,J'_{n})$ whenever two
admissible profiles $\left\langle J_{1},...,J_{n}\right\rangle $,
$\left\langle J'_{1},...,J'_{n}\right\rangle $ are permutations (reorderings)
of each other.\smallskip{}

For the next desideratum, we need a preliminary definition. For any
two evaluations $J$ and $J'$ and any subgroup $C\subseteq N$, we
say that $J$\emph{ supported by} $C$ \emph{beats} $J'$ \emph{supported
by} $N\backslash C$ if, for the profile $\left\langle J_{1},...,J_{n}\right\rangle $
in which all $i\in C$ have $J_{i}=J$ and all $i\notin C$ have $J_{i}=J'$,
the collective evaluation $F(J_{1},...,J_{n})$ is $J$ (provided
this profile is admissible). In such a profile, the set of individuals
$N$ is divided into two subgroups, one of which ($C$) supports evaluation
$J$ while the other ($N\backslash C$) supports evaluation $J'$.

\smallskip{}

\noindent\textbf{Pairwise neutrality:} The aggregation rule is neutral
between any two evaluations between which the group is divided. Formally,
for any consistent evaluations $J,J'$ and any subgroup $C\subseteq N$,
if $J$ supported by $C$ beats $J'$ supported by $N\backslash C$,
then $J'$ supported by $C$ beats $J$ supported by $N\backslash C$.\smallskip{}

\noindent\textbf{Issuewise independence:} The collective evaluation
for each issue/variable depends only on individual evaluations for
that issue/variable. Formally, for any admissible profiles $\left\langle J_{1},...,J_{n}\right\rangle $,
$\left\langle J'_{1},...,J'_{n}\right\rangle $ and any issue/variable
$x\in X$, if $J_{i}(x)=J'_{i}(x)$ for all $i\in N$, then $J(x)=J'(x)$,
where $J=F(J_{1},...,J_{n})$ and $J'=F(J'_{1},...,J'_{n})$.\smallskip{}

\noindent\textbf{Reinforcement monotonicity:} The collective evaluation
does not change if an additional individual comes to accept it as
his or her own evaluation. Formally, for any individual $i\in N$
and any admissible profile $\left\langle J_{1},...,J_{n}\right\rangle $,
if $J'_{i}=F(J_{1},...,J_{n})$, then $F(J_{1},...,J'_{i},...,J_{n})=F(J_{1},...,J_{n})$
(assuming $\left\langle J_{1},...,J'_{i},...,J_{n}\right\rangle $
is also admissible). \smallskip{}

The examples of aggregation rules mentioned earlier satisfy several,
but not all, of these desiderata. All of them \textendash{} plurality,
median, quantile, average, distance-based, and dictatorial rules \textendash{}
satisfy universality. All satisfy pairwise neutrality, with the exception
of quantile rules distinct from the median rule.\footnote{Distance-based rules satisfy pairwise neutrality, provided distance
is well-behaved.} All except the global plurality and distance-based rules satisfy
issuewise independence. All of the given rules further satisfy anonymity,
except for dictatorial ones, and all satisfy reinforcement monotonicity,
except for the averaging rule.\footnote{This includes well-behaved distance-based rules. To see that averaging
violates reinforcement monotonicity, suppose half of the individuals
assign the value 10 to $x$ and the other half 20, so the average
value is 15. If one individual now changes the evaluation for $x$
from 10 to 15, the average will change.} However, only dictatorial rules, the global plurality rule, and distance-based
rules generally satisfy collective rationality. Can we find aggregation
rules that satisfy all of the desiderata at once? We will now see
that the answer is negative whenever an evaluation problem involves
connections between three or more issues or variables. We call such
evaluation problems \emph{non-simple}.\footnote{To define \emph{non-simplicity} formally, let a \emph{partial evaluation}
be a function from some subset $X'\subseteq X$ to $V$. A partial
evaluation is \emph{consistent} if it can be extended to a consistent
evaluation in $\mathcal{J}$, and \emph{inconsistent} if not. It is
\emph{minimal inconsistent} if it is inconsistent but becomes consistent
when restricted to any proper subset of $X'$. An evaluation problem,
given by $X$, $V$, and $\mathcal{J}$, is \emph{non-simple} if it
admits a minimal inconsistent partial evaluation of more than two
issues (i.e., with $\left|X'\right|>2$). This concept generalizes
a related concept for binary judgment aggregation discussed in \cite{DietrichListArrow,DokowHolzmanBinary,NehringPuppe2010}.}

\smallskip{}

\noindent\textbf{Theorem 1:} For any non-simple evaluation problem,
there exists no aggregation rule satisfying universality, collective
rationality, anonymity, pairwise neutrality, issuewise independence,
and reinforcement monotonicity.\smallskip{}

This theorem stands in the tradition of earlier impossibility theorems
in binary and many-valued judgment-aggregation theory \cite{DietrichListArrow,DokowHolzmanBinary,DokowHolzmanNonbinary,DuddyPigginsManyValued,ListPettit2002,NehringPuppe2010,RubinsteinFishburn,Wilson}.
However, unlike earlier theorems (such as \cite{DokowHolzmanNonbinary,DuddyPigginsManyValued,DietrichRelevance}),
it applies to a wide class of realistic evaluation problems with non-binary
issues/variables and not merely to ones with highly structured interconnections
between the issues/variables. Our theorem applies, for instance, to
our macroeconomic example, where $x$, $y$, and $z$ are related
via the constraint $z=x\times y$, and to many other cases where one
variable or issue is a function of two or more others. In practice,
therefore, when a group seeks to aggregate judgments on several variables
or issues, it has no choice but to give up at least one of the six
desiderata. Arguably, universality, collective rationality, anonymity,
and pairwise neutrality are hard to drop, because they are each very
plausible and undemanding. Issuewise independence and reinforcement
monotonicity are more contestable. Why, for instance, should the collective
judgments on variable $z$ depend only on the individual judgments
on $z$, as issuewise independence demands, and not on the individual
judgments on $x$ and $y$, which seem relevant too? And why should
we rule out averaging just because it violates reinforcement monotonicity?
After all, averaging may seem independently plausible.

As we will now see, a rationale for requiring issuewise independence
and reinforcement monotonicity comes from the significance of these
two desiderata for tackling our second challenge, namely the challenge
of manipulation.

\section{The challenge of manipulation}

When we aggregate several individuals\textquoteright{} judgments on
some variables or issues into collective judgments, we would like
the individuals to have an incentive to reveal their judgments truthfully
\cite{Gibbard,Satterthwaite}. For instance, we would not like the
members of a committee or expert panel to misrepresent their judgments
for strategic purposes. Aggregation by averaging, for instance, may
incentivize some individuals to misrepresent their judgments. Suppose
an expert panel is estimating the value of some variable, and the
panel uses the averaging rule. If one of the members thinks the others\textquoteright{}
estimates are too low, this member can move the average estimate in
his or her preferred direction by reporting an estimate that is higher
than what he or she truly believes. And if the member thinks the others\textquoteright{}
estimates are too high, he or she can move the average estimate in
the preferred direction by reporting a lowered estimate. In each case,
the member will have ``manipulated'' the collective judgments by
misrepresenting his or her own judgments.

To model strategic voting (drawing on the modelling strategy in \cite{DietrichListStrategyProof,DietrichList2025},
cf. \cite{Basili}), we will say that an individual has an \emph{opportunity
to manipulate} the collective evaluation for a particular issue/variable
$x$ at a particular profile of evaluations if the following holds: 
\begin{itemize}
\item If (holding others\textquoteright{} evaluations fixed) this individual
submits his or her sincere evaluation for all issues/variables, the
collective evaluation of issue/variable $x$ will differ from the
individual\textquoteright s sincere evaluation of $x$, and 
\item if he/she suitably misrepresents his/her evaluation for some of the
issues/variables, the collective evaluation of issue/variable $x$
will move in the direction of this individual\textquoteright s sincere
evaluation of $x$. 
\end{itemize}
For the notion of ``moving in the direction of a given individual's
evaluation'' to be well-defined, we must assume that the set $V$
of values is linearly ordered, so that we can make ``less than''
(\textless ) or ``greater than'' (\textgreater ) comparisons between
different values. Formally, we now say that a given aggregation rule
$F$ is \emph{manipulable} by some individual $i\in N$ on some issue/variable
$x\in X$ at some admissible profile $\left\langle J_{1},...,J_{n}\right\rangle $
if 
\begin{itemize}
\item $J(x)\neq J_{i}(x)$, and 
\item there exists some $J_{i}'$ (where $\left\langle J_{1},...,J'_{i},...,J_{n}\right\rangle $
remains admissible) such that 
\[
J'(x)\left\{ \begin{array}{cc}
<J(x) & \text{if }J_{i}(x)<J(x),\\
>J(x) & \text{if }J_{i}(x)>J(x),
\end{array}\right.
\]
\end{itemize}
where $J=F(J_{1},...,J_{n})$ and $J'=F(J_{1},...,J'_{i},...,J_{n})$.
Averaging, for instance, is manipulable on any issue/variable $x$
by any individual $i$ who finds the average evaluation $J(x)$ too
high or too low and who can then move the average in the desired direction
by reporting an evaluation $J'_{i}(x)$ that is lower or higher than
$i$'s sincere evaluation $J_{i}(x)$.

Now the desideratum of non-manipulability demands that none of the
individuals should ever have such a manipulation opportunity.\smallskip{}

\noindent\textbf{Non-manipulability:} $F$ is not manipulable by
any individual $i\in N$ on any issue/variable $x\in X$ at any admissible
profile.\smallskip{}

Can we find non-manipulable aggregation rules? Here is where the significance
of issuewise independence and reinforcement monotonicity lies (for
related results on binary judgment aggregation or probability aggregation,
see \cite{DietrichListStrategyProof,DietrichList2025,NP2002}).\smallskip{}

\noindent\textbf{Theorem 2:} For an aggregation rule $F$ satisfying
universality, non-manipulability implies issuewise independence and
reinforcement monotonicity.\smallskip{}

In a nutshell, if a rule violates independence or reinforcement monotonicity
at a given issue, then some individual(s) can sometimes manipulate
the collective evaluation of this issue in their direction, by misrepresenting
their evaluation of that issue (if reinforcement monotonicity fails)
or by misrepresenting their evaluation of other issues (if independence
fails). The present theorem gives us necessary conditions for non-manipulability.
In supplementary material, we derive the full necessary and sufficient
conditions, which are slightly more demanding and more technical and
which we here state only informally:\smallskip{}

\noindent\textbf{Issuewise independence: }As before.\smallskip{}

\noindent\textbf{Approach monotonicity:} The collective evaluation
of an issue/variable does not change if some individual's evaluation
of that issue/variable comes closer to it.\smallskip{}

\noindent\textbf{Non-overshoot}:\textbf{ }If an individual who agrees
with the collective evaluation of an issue/variable at a given profile
unilaterally changes his or her evaluation of that issue/variable,
the collective evaluation does not move more strongly in the same
direction.\smallskip{}

The median and quantile rules, unlike the averaging rule, satisfy
all of these conditions, and so they are non-manipulable (as previously
noted for the single-issue case in \cite{BalinskiLarakiPNAS,BalinskiLaraki}).
For example, none of the participants in Galton's experiment could
have moved the median weight estimate in a preferred direction by
submitting a misrepresented weight estimate. The participants each
had the best chance of bringing about a median estimate in line with
their own estimate by revealing what they genuinely thought. Quantile
rules share this non-manipulability, but unlike the median rule, they
are not pairwise neutral. The median rule thus emerges once again
as potentially attractive. 

The observation that issuewise independence and reinforcement monotonicity
are necessary for non-manipulability, however, also heralds some bad
news. As we have seen, the median rule, which satisfies these desiderata,
does not guarantee consistent collective judgments. This problem generalizes:
if we wish to attain collective rationality and non-manipulability,
we run into an impossibility result in the tradition of earlier social-choice-theoretic
impossibility results on strategic voting \cite{Basili,DietrichListStrategyProof,DietrichList2025,Gibbard,Satterthwaite,NP2002}.\smallskip{}

\noindent\textbf{Theorem 3:} For any non-simple evaluation problem,
there exists no non-manipulable aggregation rule satisfying universality,
collective rationality, anonymity, and pairwise neutrality. \smallskip{}

This theorem is a corollary of Theorems 1 and 2. Since the last four
desiderata may be non-negotiable in many committees and decision-making
groups, it may be impossible in practice to achieve non-manipulability.
When judgments are to be aggregated on three or more issues or variables
with some constraints between them, some individuals may have an opportunity
to manipulate the collective judgments by misrepresenting their own
judgments. Whether they will always act on such an opportunity is
a different matter. This will depend on how much information these
individuals have about the judgments of others and whether they care
only about the outcome or also about their truthfulness. However,
the designers of mechanisms for aggregating judgments should be aware
of this defect of many real-world aggregation rules \textendash{}
a defect that echoes a classic impossibility result in social choice
theory, the so-called Gibbard-Satterthwaite theorem \cite{Gibbard,Satterthwaite},
and related results concerning the manipulability of binary judgment
aggregation \cite{DietrichListStrategyProof,NP2002}.

\section{The challenge of truth-tracking}

The goal is often to arrive at collective judgments that not just
reflect the individual judgments but also ``track the truth'' \cite{BovensRabinowicz,DietrichSpiekermannSEP,GrofmanCJT,ListGoodin2001,Morreau2021,Pivato}.
In Galton's example, the aim is to estimate the true weight of the
ox. Expert panels similarly seek to track the truth. In an animal
group, the aim is to navigate in the right direction, which leads
to food, safety, or a good nest-site. To analyze this demand of ``track-tracking'',
we will now suppose that, for each issue/variable, there is an independent
fact of the matter as to what its true value is. For example, there
is a fact about the true weight of the ox, the current level of atmospheric
CO$_{2}$, the true GDP under a given economic scenario, or which
travel direction leads to a desired destination. Of course, that fact
is often hard to know, but this is why we require groups of individuals
to judge what it is. 

To model the truth-tracking performance of different aggregation rules,
we will consider potentially infinitely many individuals $i=1,2,3,...$
from whom different-sized groups of evaluators could be drawn. For
any number $n$, the first $n$ individuals will then form a group
of size $n$. We retain our earlier definitions of the set $X$ of
variables or issues under consideration, the set $V$ of possible
values, and the consistency constraints encoded by $\mathcal{J}$.
However, we now define an \emph{aggregation rule} as a function that
maps any admissible profile of evaluations $\left\langle J_{1},...,J_{n}\right\rangle $
\emph{of any length} $n$ to a collective evaluation $J=F(J_{1},...,J_{n})$.
Technically, $F$ has the ``variable-group-size universal domain'',
$\bigcup_{n=1}^{\infty}\mathcal{J}^{n}.$ Our earlier examples of
aggregation rules can easily be redefined in this variable-group-size
way.\footnote{For aggregation rules that are defined only for an odd group size
$n$, like the median rule, the variable-group-size universal domain
must be defined as $\bigcup_{n=1,3,5,...}\mathcal{J}^{n}.$} 

We model the facts to be tracked and the judgments about these facts
as random variables relative to some underlying probability space
with probability measure $P$. The facts can be modelled as a random
variable $\mathbf{S}$, which we will call the \emph{situation} and
which encodes the factually true values of the issues or variables
in $X$. As the situation fixes the true evaluation of those issues/variables,
we assume that $\mathbf{S}$ takes values from the set $\mathcal{J}$.
For each issue/variable $x$, $\mathbf{S}(x)$ represents the true
value of $x$. (Later we will generalize the definition of $\mathbf{S}$.)
We further assume that every value $S$ taken by $\mathbf{S}$ \textendash{}
for short, every \emph{situation} \textendash{} has positive probability,
i.e., $P(S)>0$. The individual evaluations can be modelled as random
variables too, namely $\mathbf{J}_{1},\mathbf{J}_{2},\mathbf{J}_{3},...$,
taking values in $\mathcal{J}$. 

This model allows us to represent, for instance, the probability that
individual $i$'s judgment about the situation is correct, $P($$\mathbf{J}_{i}=\mathbf{S})$,
or the probability that $i$'s judgment on issue or variable $x$
is too low or too high, $P($$\mathbf{J}_{i}(x)<\mathbf{S}(x))$ and
$P($$\mathbf{J}_{i}(x)>\mathbf{S}(x))$. We thus have a framework
for investigating the probability of obtaining correct collective
judgments under various aggregation rules and various scenarios.

For our first baseline result, we make the idealizing assumptions
that the individuals are\emph{ independent}, \emph{judgmentally homogeneous},
and \emph{competent} on every issue/variable. \emph{Independence}
means that the random variables $\mathbf{J}_{1},\mathbf{J}_{2},...$
are probabilistically independent, given the situation $\mathbf{S}$.
\emph{Judgmental homogeneity} means that these random variables are
identically distributed, given \textbf{$\mathbf{S}$}. \emph{Competence}
on an issue/variable $x$ means that each individual $i$ has a probability
greater than $\frac{1}{2}$ of not underestimating the value of $x$
and also a probability greater than $\frac{1}{2}$ of not overestimating
that value, regardless of the situation that obtains.\footnote{Formally, we define individual $i$'s \emph{non-underestimation competence}
on issue/variable $x$ as $\mathbf{C}_{i,x}^{\geq}=P(\mathbf{J}_{i}(x)\geq\mathbf{S}(x)\vert\mathbf{S})$
and $i$'s \emph{non-overestimation competence} on $x$ as $\mathbf{C}_{i,x}^{\leq}=P(\mathbf{J}_{i}(x)\leq\mathbf{S}(x)\vert\mathbf{S})$.
Both are random variables, because the probability of not underestimating
or overestimating the value of a given issue/variable can vary from
situation to situation. \emph{Competence} on issue/variable $x$ then
means that $\mathbf{C}_{i,x}^{\geq}>\frac{1}{2}$ and $\mathbf{C}_{i,x}^{\leq}>\frac{1}{2}$,
regardless of the situation.} In an idealized version of Galton's experiment, the participants
estimating the weight of the ox might be independent, judgmentally
homogeneous, and competent. They each have a probability greater than
$\frac{1}{2}$ of not overestimating the weight of the ox, and also
a probability greater than $\frac{1}{2}$ of not underestimating it.
Note that this permits a low probability of judging the weight with
perfect accuracy. Under these idealized assumptions, the following
holds: \smallskip{}

\noindent\textbf{Theorem 4:} If individuals are independent, judgmentally
homogeneous, and competent on every issue/variable, the median rule
is increasingly accurate and asymptotically infallible on every issue/variable.\smallskip{}

Here, being\emph{ increasingly accurate} on some issue/variable $x$
means that the probability that the aggregation rule evaluates this
issue correctly increases with increasing group size $n$;\footnote{Formally, this probability is $P(F(\mathbf{J}_{1},...,\mathbf{J}_{n})(x)=\mathbf{S}(x))$.
The increase is \emph{weak}, meaning that the probability either increases
or remains equal as $n$ goes up. Furthermore, to avoid complications
due to ties, the group size $n$ is restricted to odd numbers (1,3,5,...).} and being \emph{asymptotically infallible} on some issue/variable
$x$ means that this probability converges to 1 as $n$ tends to infinity.
Mathematically, the present theorem is a multi-issue version of a
jury theorem for grade aggregation, where different evaluators each
assign grades such as A, B, C etc. to a given option and the collective
grade is defined as the median grade \cite{Morreau2021}.\footnote{The latter result (from \cite{Morreau2021}) can be extended to quantile
rules, which outperform the median rule when individuals have an upside
or downside bias in their evaluations \cite{Kilinc2025}. In \cite{Kilinc2025},
individuals' non-overestimation and non-underestimation competence
need not both exceed 1/2, as long as their sum exceeds 1. The median
rule must then be replaced by a $p$-quantile rule with $p$ calibrated
so as to compensate for any upside or downside bias in individual
evaluations. Our present theorem could be analogously extended to
quantile rules, so as to allow for individual evaluation biases.} 

Theorem 4 can be viewed as a generalization of the classic ``jury
theorem'' that goes back to Nicolas de Condorcet in the 18th century
\cite{GrofmanCJT,ListGoodin2001,DietrichSpiekermannSEP}. Condorcet
had shown that, in a collective decision on a single yes/no question
(such as whether a defendant in a trial is guilty) with decision-makers
who make correct judgments with independent and identical probabilities
greater than $\frac{1}{2}$, the probability of a correct majority
judgment increases and converges to 1 as the group size increases.
Our result extends this to non-binary judgments (in line with \cite{Morreau2021})
and to any number of issues/variables. While we focus on the median
rule here, we briefly address the truth-tracking properties of averaging
in online supplementary material. (Recall again that averaging has
several disadvantages compared to the median rule: it rests on more
demanding measurability assumptions, is not robust to outlier judgments,
and violates non-manipulability.)

A consequence of Theorem 4 is that, under its idealized assumptions,
the probability of an inconsistency in the collective judgments made
by the median rule (i.e., a breach of the constraints encoded by $\mathcal{J}$)
converges to 0 as $n$ tends to infinity, because the truth that is
being accurately tracked is consistent. So, the median rule seems
to be able to address \emph{all} three challenges we have discussed,
at least asymptotically: the challenges of collective inconsistency,
manipulation, and truth-tracking.

Unfortunately, however, the assumptions of Theorem 4 are very unrealistic.
To begin with, the assumption of judgmental homogeneity is unrealistic.
If there is heterogeneity across the individuals, they may not all
be equally competent. This limitation, however, can still be addressed
relatively easily. The conclusion of increasing accuracy in Theorem
4 can be restored by introducing randomness about the identity of
additional group members; and the conclusion of asymptotic infallibility
continues to hold under the assumption that individuals are competent
\emph{on average} on every issue/variable \emph{x}, i.e., that the
average probability of not underestimating the value of $x$ and the
average probability of not overestimating it both exceed $\frac{1}{2}$,
regardless of the situation that obtains.\footnote{Using the formalism from a previous footnote, the average non-underestimation
competence on issue/variable $x$ in a given group of size $n$ is
$\frac{1}{n}\sum_{i=1}^{n}\mathbf{C}_{i,x}^{\geq}$, and the average
non-overestimation competence is $\frac{1}{n}\sum_{i=1}^{n}\mathbf{C}_{i,x}^{\leq}$.
\emph{Competence on average} on $x$ means that $\frac{1}{n}\sum_{i=1}^{n}\mathbf{C}_{i,x}^{\geq}$
and $\frac{1}{n}\sum_{i=1}^{n}\mathbf{C}_{i,x}^{\leq}$ each exceed
$\frac{1}{2}$ in the limit (or more generally limit inferior) as
$n\rightarrow\infty$, regardless of the value of $\mathbf{S}$.} We can thus treat judgmental homogeneity as a convenient simplifying
assumption. 

The other limitations of Theorem 4, however, are more serious (for
reasons similiar to those discussed in \cite{DietrichList2004,DietrichSpiekermann2013,DietrichSpiekermannSEP}).
First of all, different individuals' judgments are not normally independent
from each other in the strong sense that, holding fixed the true values
of the relevant issues or variables, there are no probabilistic dependencies
between them. The individuals may have communicated, relied on shared
or partially shared evidence, and be affected by shared environmental
conditions. In consequence, there may be common causes affecting all
of their judgments. Secondly, the assumption of issuewise competence
is unrealistic, even if we make the simplifying assumption of homogeneity.
If the situation includes common causes and other factors influencing
the individual judgments, it is unrealistic to assume that \emph{regardless
of the situation} each individual \emph{always} has a probability
greater than $\frac{1}{2}$ of not underestimating the value of any
issue/variable and also a probability greater than $\frac{1}{2}$
of not overestimating it. The evidence or other factors may have a
misleading influence and make it sometimes more likely that individuals
under- or overestimate the value of some issues or variables. 

To address those limitations, we must amend our model. First, we must
specify the situation $\mathbf{S}$ more richly, so as to include
not only the true values of the variables or issues but also other
possible common causes, from shared deliberation and shared evidence
to any other shared influences that may compromise independence in
the original sense. Formally, we introduce hypothetical issues or
variables beyond those in the set $X$ (building on the modelling
strategy in \cite{DietrichSpiekermann2013,DietrichSpiekermannSEP}).
So, we now introduce a possibly larger set $X^{+}$ of issues or variables
(a superset of $X$) and a possibly larger set $V^{+}$ of values
(a superset of $V$) and define a \emph{value configuration} to be
a function from $X^{+}$ to $V^{+}$. We write $\mathcal{J}^{+}$
to denote the (non-empty) set of all value configurations that are
deemed possible. We assume that a value configuration is possible
(i.e., in $\mathcal{J}^{+}$) only if its restriction to the issues
or variables in $X$ is consistent (i.e., in $\mathcal{J}$). The
\emph{situation} in this generalized sense is a random variable $\mathbf{S}$
with values in $\mathcal{J}^{+}$.\footnote{Again, we assume that $P(S)>0$ for each situation $S$, i.e., each
value of $\mathbf{S}$. What we previously defined is the special
case where $X=X^{+}$, $V=V^{+}$, and $\mathcal{J}=\mathcal{J}^{+}$.} With this generalization in place, independence becomes plausible
again: if we conditionalize on all common causes of the different
individuals' judgments, there are no further dependencies between
them.

Secondly, we must capture the fact that some situations may be misleading
in the sense that they make it more likely that individuals over-
or underestimate the value of some issues or variables. In what follows,
we require only a \emph{tendency} for individuals to be competent
on those issues or variables, not a \emph{guaranteed} competence \cite{DietrichSpiekermann2013,DietrichSpiekermannSEP}.
The important point is that the individuals' probabilities of not
overestimating or not underestimating the value of a given issue or
variable \textendash{} for short, their ``competence levels'' \textendash{}
are not constant but random variables that depend on the situation.
In some situations, competence levels may be higher, in others lower.
We say that individual $i$ \emph{tends to be competent} on an issue
or variable $x$ if $i$'s probability of not underestimating the
value of $x$ \emph{tends to exceed} $\frac{1}{2}$ and $i$'s probability
of not overestimating that value also \emph{tends to exceed} $\frac{1}{2}$.
Formally, this means that each of these probabilities (``competence
levels'') takes the value $\frac{1}{2}+\epsilon$ with at least as
high a probability as the value $\frac{1}{2}-\epsilon$, for every
$\epsilon>0$. Figure 1 (adapted from \cite{DietrichSpiekermannSEP})
shows an example of what such a probability distribution over competence
levels might look like. In this example, nine competence levels are
possible (0, 0.125, 0.25, 0.375, 0.5, 0.625, 0.75, 0.875, 1), and
each competence level above $\frac{1}{2}$ is more probable than the
opposite level below $\frac{1}{2}$.
\begin{figure}[h]
\centering{}\caption{Competence as a random variable}
\includegraphics[scale=0.4]{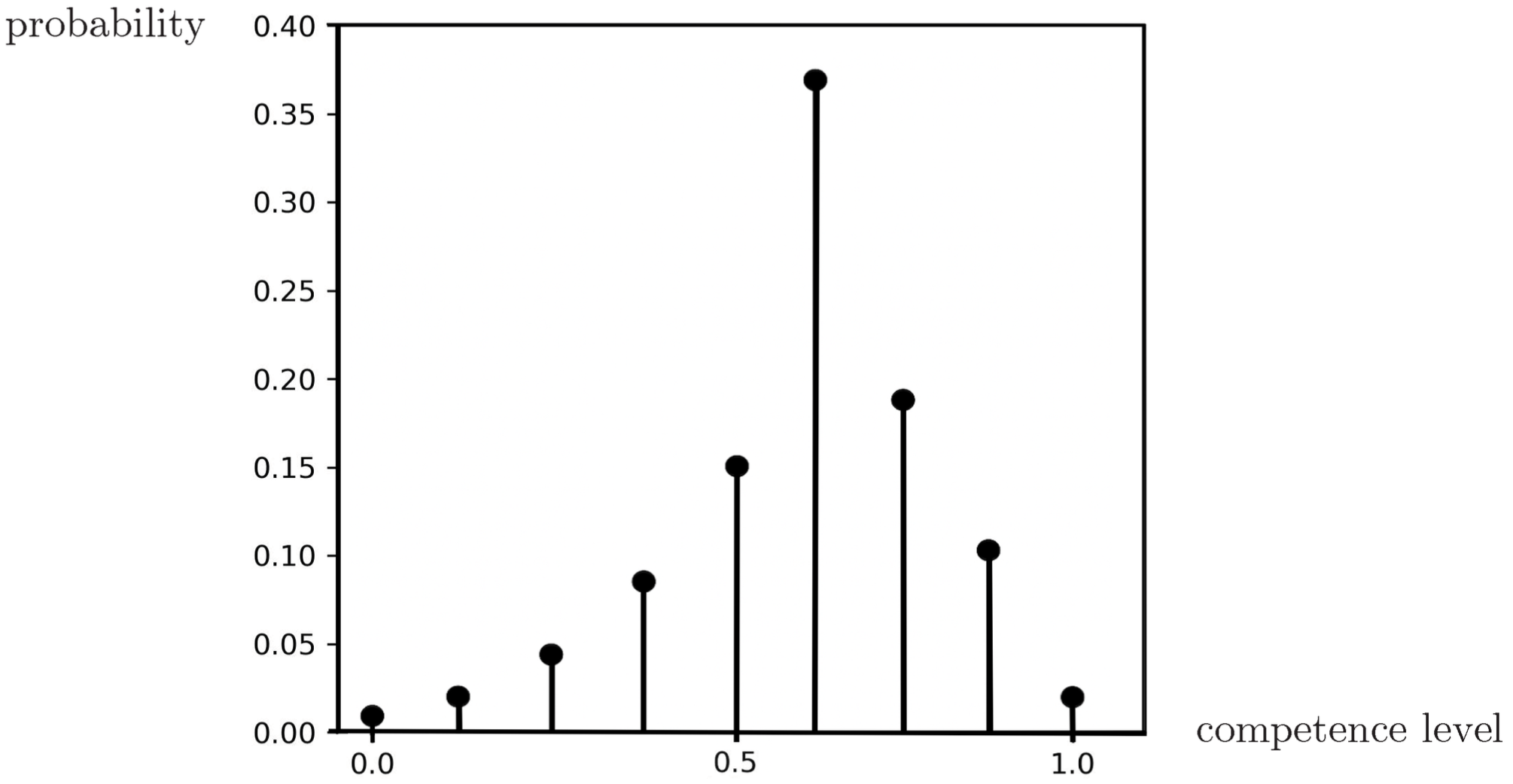}
\end{figure}
 A \emph{tendency to be competent} thus permits the occurrence of
situations in which the individuals' competence levels are less than
$\frac{1}{2}$. In Figure 1, for instance, it can easily happen that
individual competence is only 0.375 or even lower. But such a competence
level is less likely than the opposite competence level of 0.625 or
higher. The previous definition of ``guaranteed'' competence is
the special case where competence levels are \emph{always} above $\frac{1}{2}$.\smallskip{}

\noindent\textbf{Theorem 5:} If individuals are independent and judgmentally
homogeneous, the median rule is increasingly accurate on any issue/variable
on which individuals tend to be competent, and asymptotically fallible
on it unless individuals are always competent on it.\smallskip{}

This theorem (which generalizes a result from \cite{DietrichSpiekermann2013}
to non-binary evaluations) captures a less unrealistic form of ``collective
intelligence''. First, the theorem's independence assumption is stated
against the background of a sufficiently rich specification of the
situation, which can include common causes of the individual judgments.
Second, the individuals only need to have a tendency to be competent
\textendash{} and only on certain issues or variables. The median
rule performs well on those issues, in the sense that larger groups
of decision-makers are more reliable than smaller groups. But we can
no longer derive asymptotic infallibility; the collective reliability
is typically bounded above by some threshold below 1, due to a non-zero
probability that the situation is misleading. Crucially, however,
the median rule retains its key merits when restricted to issues or
variables on which individuals tend to be competent.

\section{Concluding remarks}

Groups must often aggregate judgments on non-binary issues or variables.
The problem arises not only in human decision-making committees or
expert panels, but also in animal groups choosing one option among
many: travel directions, grazing spots, or nest sites. 

It is surprisingly difficult to find an aggregation rule that simultaneously
addresses the challenges of inconsistency, manipulation, and truth-tracking.
While the challenges of inconsistency and manipulation are arguably
most relevant to the human case, the truth-tracking challenge is highly
relevant for animal group decisions too.\footnote{The inconsistency challenge is less relevant to non-human animals
to the extent that animal group decisions do not involve multiple
interconnected issues. It is an intriguing question whether there
could be ``strategic voting'' in animal groups that use a manipulable
aggregation rule, such as averaging.} Overall, the median rule performs reasonably well: it is non-manipulable
and has increasing truth-tracking accuracy on those issues/variables
on which individuals tend to be competent, such as the weight of an
ox or the best travel route to some destination. But it does not generally
overcome the challenge of inconsistency, except asymptotically when
the unrealistic assumptions of our initial Condorcet-inspired jury
theorem are met.

How, then, can we achieve collective intelligence through aggregation?
We briefly consider two proposals that retain some of the median rule's
merits while avoiding inconsistencies across different issues. One
proposal is to drop the universality requirement and to restrict the
domain of admissible inputs of the median rule to those profiles $\left\langle J_{1},...,J_{n}\right\rangle $
that have the special property of \emph{comonotonicity} with respect
to the given linear order on $V$: for all individuals $i,j$, either
(i) for every $x\in X$, $J_{i}(x)\leq J_{j}(x)$, or (ii) for every
$x\in X$, $J_{i}(x)\geq J_{j}(x)$. Comonotonicity means that, for
any two individuals $i$ and $j$, the move from $i$'s to $j$'s
evaluation goes in the same direction for every issue. When restricted
to comonotonic profiles of consistent evaluations, the median rule
(for odd $n$) satisfies collective rationality, while still satisfying
all the other desiderata from Theorem 1. The reason is that, for any
comonotonic profile $\left\langle J_{1},...,J_{n}\right\rangle $,
there exists an individual $i$ whose evaluation $J_{i}$ lies in
the middle on \emph{every} issue, and the median rule selects this
evaluation, $J_{i}$, which is consistent. For a similar reason, any
quantile rule is also consistent for comonotonic profiles, but of
course not generally pairwise neutral.\footnote{When all issues/variables are binary, comonotonicity becomes a special
case of a known property called ``unidimensional alignment'' \cite{List2003}.
The full generalization of unidimensional alignment in the multi-valued
case is more complicated. A profile $\left\langle J_{1},...,J_{n}\right\rangle $
is \emph{unidimensionally aligned} if the individuals can be linearly
ordered on an axis from ``left'' to ``right'' such that, for each
issue/variable $x\in X$, \emph{either} $J_{i}(x)$ weakly increases
as $i$ moves from left to right \emph{or} $J_{i}(x)$ weakly decreases
as $i$ moves from left to right. Comonotonicity is the special case
that leaves out one of the two disjuncts. For odd $n$, the median
rule is also consistent for every unidimensionally aligned profile
$\left\langle J_{1},...,J_{n}\right\rangle $. By contrast, other
quantile rules are generally consistent only for comonotonic profiles.
We thank a referee for prompting us to consider comonotonicity as
an escape route from the impossibility result of Theorem 1.}

A second proposal is to give up issuewise independence and to aggregate
judgments in a ``premise-based way'' (building on similar ideas
in binary judgment and probability aggregation; e.g., \cite{BovensRabinowicz,DietrichList2017,DietrichMongin,KornhauserSager1986,ListPettit2002}).
A \emph{premise-based aggregation rule} works as follows:
\begin{enumerate}
\item We designate some issues/variables as ``premises'', such as $x$
(productivity) and $y$ (employment level) in our earlier example.
\item We aggregate the judgments for those issues/variables via the median
criterion.
\item We use some relevant inference rules, such as the constraint $z=x\times y$,
to derive the collective judgments for other issues/variables, such
as $z$ (GDP). 
\end{enumerate}
The resulting judgments will always be consistent. This approach seems
plausible in particular when the issues/variables can be subdivided
into independent and dependent variables (or ``basic'' and ``derived''
variables) and when the judgments on the former are given priority
over those on the latter. The idea is that the values of the dependent
(``derived'') variables are a function of the values of the independent
(``basic'') variables (the ``premises''), and we can obtain collective
estimates of the dependent variables by aggregating the inividuals'
estimates of the independent variables and then applying the relevant
function to derive the resulting values of the dependent variables.

The premise-based approach makes the collective judgments non-manipulable
on the ``premise'' variables, and it has a reasonable truth-tracking
performance on them as long as the individual evaluators tend to be
competent on them. The downside is that the collective judgments will
be sensitive to the choice of the premise issues or variables. This
may be no problem when there is a clear criterion for distinguishing
between premises and non-premises, for instance when it is clear which
variables count as independent and which count as dependent. But if
we are dealing with interconnected variables in a complex system,
such as the economy or the climate, such a clearcut criterion may
not exist. Decision-making groups should be aware of the challenges
identified here.

Finally, our analysis can accommodate two distinct cases of many-valued
judgments: first, cases where the issues/variables to be estimated
are non-binary (such as the weight of an ox, the level of atmospheric
CO$_{2}$, the travel direction); and second, cases where the issues/variables
themselves are fundamentally binary (such as whether a particular
scientific hypothesis is true or whether a particular event will happen)
but the evaluators' \emph{attitudes} towards them are non-binary (such
as degrees of belief or degrees of desire or approval). Although our
discussion has focused mostly on the first case, our model and results
can equally accommodate the second. This flexibility, along with the
ubiquity of general aggregation problems in many decision-making groups
\textendash{} human and non-human \textendash{} should make our analysis
widely applicable.

\end{document}